\documentclass[conference]{IEEEtran}
\IEEEoverridecommandlockouts
\usepackage{cite}
\usepackage{amsmath,amssymb,amsfonts}
\usepackage{algorithmic}
\usepackage{graphicx}
\usepackage{textcomp}
\usepackage{xcolor}
\def\BibTeX{{\rm B\kern-.05em{\sc i\kern-.025em b}\kern-.08em
    T\kern-.1667em\lower.7ex\hbox{E}\kern-.125emX}}

\usepackage[utf8]{inputenc}
\usepackage[T1]{fontenc}

\usepackage{hyperref}

\usepackage{microtype}
\usepackage{listings}
\definecolor{commentcolor}{rgb}{0.5,0.5,0.5}
\definecolor{darkgreen}{rgb}{0.09, 0.45, 0.27}
\definecolor{bgcolor}{rgb}{0.99,0.99,0.99}

\newcommand{\lstsize}{\footnotesize}

\def\backtick{\`{}}
\lstdefinestyle{prompt}{
    breakatwhitespace=true, 
    columns=fullflexible,
    tabsize=2,%
    breaklines=true,
    keywordstyle=\color{blue},
    escapeinside={|}{|},
    breakindent=0pt,
    frame=single,
    literate={`}{\backtick}1,
}
\lstdefinelanguage{dsl}{
    sensitive=true,      
    morekeywords={ensemble,component,components,attribute,description,format,beyond,control,name,assign,as,into,ensembles,singleton,foreach,strategy,interface,class,answer_format,chain_of_thought,auc_interface,am_interface,global,if,param,extends,beyond-control,in,for,periodically,constraint,conversation_history,max_retries,reason,context_summary,situation,initial,state},
    alsoletter={-},
    morestring=[b]",
    morecomment=[l]{\#},
    keywordstyle=\color{blue},
    stringstyle=\color{darkgreen},
    showstringspaces=false,
    commentstyle=\color{gray},
}
\lstdefinestyle{dsl}{
    basicstyle=\linespread{0.85}\sffamily\lstsize,
    language=dsl,
    breakatwhitespace=true, 
    columns=fullflexible,
    tabsize=2,%
    breaklines=true,
    escapeinside={|}{|},
    breakindent=0pt,
    frame=single,
}

\newcommand{\MAX}[0]{\mathit{MAX}}
\newcommand{\INF}[0]{\mathit{INF}}
\newcommand{\BEG}[0]{\mathit{BEG}}

\usepackage{scrextend}
\newcommand{\replicationFootnote}{\footnote{\label{replication}
\url{https://github.com/smartarch/llm-adaptation/tree/icsa}
}}
\newcommand{\replication}{\footref{replication}}

\newcommand{\blue}[1]{\textcolor{blue}{#1}}
\newcommand{\green}[1]{\textcolor{darkgreen}{#1}}

\newcommand{\violet}[1]{\textcolor{violet}{#1}}

\usepackage{subcaption}
\usepackage{footmisc}
\usepackage{tabularx,colortbl}
\usepackage{array}
\usepackage[inline]{enumitem}
\usepackage{makecell}
\usepackage{multirow}

\begin{document}
\title{Feedback-based Automated Verification in Vibe Coding of CAS Adaptation Built on Constraint Logic}

\author{%
	\IEEEauthorblockN{%
	    Michal Töpfer,
        František Plášil,
        Tomáš Bureš,
        Petr Hnětynka
	}%
	\IEEEauthorblockA{%
        \textit{Charles University, Czech Republic}\\
        Email: \{michal.topfer, frantisek.plasil, tomas.bures, petr.hnetynka\}@matfyz.cuni.cz
    }%
}

\maketitle

\begin{abstract} In CAS adaptation, a challenge is to define the dynamic architecture of the system and changes in its behavior. Implementation-wise, this is projected into an adaptation mechanism, typically realized as an Adaptation Manager (AM).
With the advances of generative LLMs, generating AM code based on system specification and desired AM behavior (partially in natural language) is a tempting opportunity. 
The recent introduction of vibe coding suggests a way to target the problem of the correctness of generated code by iterative testing and vibe coding feedback loops instead of direct code inspection. 

In this paper, we show that generating an AM via vibe coding feedback loops is a viable option when the verification of the generated AM is based on a very precise formulation of the functional requirements. We specify these as constraints in a novel temporal logic FCL that allows us to express the behavior of traces with much finer granularity than classical LTL enables.

Furthermore, we show that by combining the adaptation and vibe coding feedback loops where the FCL constraints are evaluated for the current system state, we achieved good results in the experiments with generating AMs for two example systems from the CAS domain. Typically, just a few feedback loop iterations were necessary, each feeding the LLM with reports describing detailed violations of the constraints. This AM testing was combined with high run path coverage achieved by different initial settings. 

\end{abstract}


\section{Introduction}
\label{sec:introduction}
Collective Adaptive Systems (CAS) feature cooperating entities (frequently called agents) collaborating in groups (ensembles) towards a common goal. As the name suggests, adaptation is an integral feature of CAS. Adaptation concerns the dynamic presence of agents, their flexible grouping, and emerging changes in the environment. This inherently means dynamic modification and adaptation of the system architecture. Obviously, a challenge is to define the dynamic system architecture and changes in the behavior of the system.  To this end, \cite{Denicola} provides an overview of a number of different approaches covering, e.g., goal-oriented design, adaptation feedback loops, logic-based modeling and analysis, and attribute and aggregate programming techniques. Overall, CAS adaptation is a challenging task where the particular domain can play a substantial role so that its realization has to reflect the application specifics. Implementation-wise, this has to be projected in the adaptation mechanism (typically realized as a software block \textit{Adaptation Manager} -- AM later on). 

With the advances of generative LLMs and the vision of employing generative AI in software architecture~\cite{Muccini}, the option to generate code based on providing a prompt containing the system specification (even partially in natural language) is a tempting opportunity for generating an AM. As for every code generated this way, the correctness of the result is a key issue, so human code inspection originally appeared as a necessity.  For example, \cite{Tpfer2025}~reports underwhelming results when experimenting with generation of AM code per se, i.e., without any feedback to LLM. 

Nevertheless, the recent introduction of vibe coding suggests a way around the problem (a comprehensive survey is, e.g., in~\cite{ge2025surveyvibecodinglarge}). Instead of direct code inspection, vibe coding proposes its checking in iterative feedback loops and asking the LLM for its improvement by prompt adjustments. Such prompt modification may potentially include human interventions as well (for instance, to enhance/correct the system specification). Code checking can be based on a wide range of methods~\cite{ge2025surveyvibecodinglarge}; for example, in~\cite{inproceedings} it ranges from handling compilation errors, running predefined tests, static analysis, employing LLM-generated tests, to injecting flaws into the code to assess the quality of the tests.   
In this paper, we aim at targeting the challenge of to which extent to combine a formal way of system specification and its description in a natural language to achieve a viable result in vibe coding an AM.

Instead of considering the whole spectrum of code assessment methods, we further particularly focus on testing the AM code for violation of a set of architectural constraints. These include \textit{generic constraints} that ensure compliance with general architectural rules (e.g., each component is assigned to exactly one ensemble, i.e., a collaborating group of components), and \textit{functional constraints} that ensure that each architectural adaptation step will preserve the desired functionality of the particular application. We propose an adaptation specification in a DSL that combines a description of the high-level adaptation strategy in natural language and a formalization of the detailed constraints that the AM should follow and comply with. To this end, we designed and formalized a novel temporal logic -- Functional Constraints Logic (FCL), allowing the definition of adaptation steps' constraints in both a detailed and an easy-to-reflect-in-DSL way. 

In this context, we focus on the following research questions.
\begin{enumerate}[leftmargin=3.2em,start=1,label={\bfseries RQ\arabic*:},ref={\bfseries RQ\arabic*},nosep,topsep=2pt]
\item\label{RQ1} How to automatically guide generative AI in designing and implementing CAS adaptation as an AM based on system specification via vibe coding?
\item\label{RQ2} How to ensure that vibe coding produces AM yielding valid adaptations (i.e., conforming to the generic and functional constraints)?
\item\label{RQ3} How efficient is the method of combining the vibe coding feedback loop and the adaptation loop?
\end{enumerate}

The contributions of this paper include
\begin{itemize}
    \item design of a DSL for specifying the system and its architecture (ADSL),
    \item design of a novel temporal logic FCL as a base for the specification of functional constraints in a dedicated DSL (FCDSL),
    \item a tool for runtime verification of an AM, including evaluation of the functional constraints,
    \item an experimentation framework for generating LLM prompts from architectural specification, querying the LLMs via an API, and providing feedback to the LLM based on runtime verification,
    \item evaluation on two example cases.
\end{itemize}
The implementation of the framework, experimental setup, and raw results are available in the replication package\replicationFootnote.

The structure of the paper is as follows. 
In \autoref{sec:background}, we introduce the running example and show how to model it as CAS. \autoref{sec:approach} describes the approach of combining vibe coding with runtime verification feedback to implement an AM, while \autoref{sec:constraints} gives more details on FCL and defining functional constraints.
In \autoref{sec:evaluation}, we evaluate our approach and provide the results of the experiments. \autoref{sec:related} discusses related work and \autoref{sec:conclusion} concludes the paper.

\section{Background and Running Example}
\label{sec:background}
\subsection{Running example}
\label{sec:runningexample}

As a running example, we use a simple made-up ``Dragon Hunt'' game, 
adopted from \cite{Tpfer2025}.
This example is intentionally fictional, so that the LLM could not have been trained on its exact rules.

A Dragon lives in a Cave near a Village. The goal of the game is to kill the Dragon. 
The player of the game can control the villagers by assigning them tasks (in each of his/her turns).
Villagers can either stay in the Village and produce wheat, that allows them to spawn more villagers, or move to the Cave to attack the Dragon. The Dragon attacks back, possibly killing some of the villagers.
There is a limit of 30 turns (steps), after which the player loses if the Dragon remains alive.

The rules of the game define what tasks can be assigned to the villagers. Villagers in the Village can
\begin{enumerate*}
    \item farm (produce wheat),
    \item spawn a new villager,
    \item go to the Cave.
\end{enumerate*}
Villagers in the Cave can
\begin{enumerate*}
    \item \label{a} attack the Dragon,
    \item \label{b} stay idle in the Cave,
    \item \label{c} go back to the Village.
\end{enumerate*}
There are two roles of villagers in the game: farmers (more efficient at farming wheat) and warriors (capable of more damage to the Dragon).

\subsection{Modeling the running example as a CAS}

To model the running example as a CAS, we employ an ensemble-based software architecture based on the concepts from the DEECo model~\cite{bures_deeco_2013,bures_language_2020}.

The villages are the \textit{components} of the system. Each component has several attributes representing the component's state (location, health points, role, etc.). Furthermore, there are \textit{beyond-control components} that cannot be controlled by the AM, but their attributes can be observed, such as the current amount of wheat in a silo in the Village.

The system is adapted in discrete time steps (turns of the game). In each step, the AM performs \textit{ensemble resolution} -- it divides the components into groups (called \textit{ensembles}). This updates the architecture of the system (the composition of components) and assigns tasks to components.
For example, all the components assigned to the ensemble \lstinline{Farm} will work on the farm to produce wheat.

To make the assignment of tasks to components simpler, we require that during ensemble resolution each component is assigned to exactly one ensemble.
The rules of the game also constrain the architecture, e.g., only the components located in the Village can be assigned to the \lstinline{Farm} ensemble.

From a broader perspective, the division of components into ensembles is an instance of the grouping problem, being NP-hard in general \cite{KOVALYOV20101908}. Therefore, a viable way around such complexity is to find a suitable heuristic -- an opportunity to ask LLM to propose one and find a way to check that what it does propose meets the specification of the problem in question. 

From the perspective of the MAPE-K loop~\cite{mapek} commonly used as a model of the adaptation process, the AM observes the state of the system (Monitor), decides on the ensemble resolution (Analyze + Plan), and assigns tasks to the components based on their membership in a particular ensemble (Execute).

\section{Vibe Coding and Feedback Verification}
\label{sec:approach}
\subsection{Combining adaptation and feedback loops}
\label{sec: combining adaptation and feedback loops}
To answer \ref{RQ1}, we propose combining vibe coding with an automated feedback loop to the LLM based on the results of runtime verification done as part of the adaptation loop.
It consists of two main stages executed in these loops:
\begin{enumerate}
    \item Prompting an LLM to generate the code of the AM (\autoref{sec:approach:prompt-generation}),
    \item Runtime verification of the generated AM (\autoref{sec:approach:verifier}) that potentially yields feedback to code generation.
\end{enumerate}

\begin{figure}[tbh]
	\includegraphics[width=\linewidth]{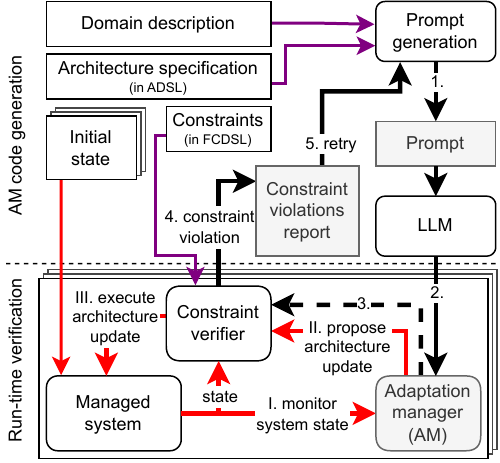}
	\caption{Combining adaptation and feedback loops.}
	\label{fig:approach}
\end{figure}

This idea is graphically expressed in \autoref{fig:approach} and is further elaborated in this section.

At design time, the architect specifies the goals of the system and constraints on the desired architecture of the components and ensembles and their state. From this specification, a prompt is automatically generated (``1.'' in \autoref{fig:approach}) and presented to the LLM. The task of the LLM is to design and implement an AM  for the system -- it should output it as a block of Python code (``2.'').

The system is run, and the generated AM periodically adapts the system architecture (red loop in \autoref{fig:approach}):
It reads the monitored system state (``I.'') and proposes an update to the system architecture (``II.''). The architecture update is executed on the system (``III.'') and the system runs until the next adaptation (``I.'' again).

At the same time, run-time verification of the functionality of AM takes place (``3.'').  Its important part is the constraint verifier that checks whether predefined constraints are not violated, in particular:
\begin{itemize}
    \item \textit{Generic constraints} ensure that the AM code complies with key specification requirements imposed for the adaptation loop environment, and that the proposed architecture updates meet the generic rules of the component and ensemble cooperation.
    More details on these constraints are given in \autoref{sec:approach:verifier}.
    \item \textit{Functional constraints} ensure that the AM follows the adaptation strategy specified in ADLS (\autoref{lst:dsl-dragon}). These constraints are described in more detail in \autoref{sec:approach:verifier} and \autoref{sec:constraints}.
\end{itemize}
If any constraints are violated, they are adequately recorded in the constraint violation report (``4.''), the system run is stopped and the feedback loop is activated -- black loop in \autoref{fig:approach}.

In particular, the LLM is asked to generate the AM again to fix these violations (``5.''). The constraint violation report is made a part of the new prompt (the original prompt and the LLM response are kept in the context). This closes the feedback loop. The updated AM is again verified and the loop is repeated until an AM passing all the constraints is generated (a safety mechanism can also be employed to stop the feedback loop after a predefined number of unsuccessful iterations).

\subsection{Prompt generation}
\label{sec:approach:prompt-generation}

The prompt to the LLM is automatically generated from a \textit{domain description} (\autoref{sec:approach:domain}) and an \textit{architecture specification} (in the domain specific language ADSL; \autoref{sec:approach:architecture}). \autoref{sec:approach:prompt} overviews how the prompt is generated.

\subsubsection{Domain description}
\label{sec:approach:domain}

The domain description defines the system scenario in natural language. It describes the rules of the system, specifies its entities, behavior requirements, etc. Domain description of the running example is illustrated in \autoref{lst:dragon-domain}.

\begin{lstlisting}[style=prompt,label=lst:dragon-domain,caption=Domain description for the Dragon Hunt.,xleftmargin=10pt,breakindent=-10pt,framexleftmargin=10pt]
You play a computer game. The goal is to kill a Dragon as fast as possible.
The Dragon lives in a Cave near a Village. The player controls villagers. The villagers can either be in the Village and work on a farm (and produce wheat) or they can go to the Cave, where they can attack the Dragon.
There are two types of villagers - farmers and warriors. Farmers produce more wheat when they work on a farm, and warriors deal more damage to the Dragon. Farmers start with 4 HP, produce 5 wheat when farming, and deal 1 damage when attacking. Warriors start with 6 HP, produce 2 wheat when farming, and deal 3 damage when attacking.
To spawn a new villager, at least two villagers must be in the "spawn" group and some wheat is consumed (12 for warriors, 10 for farmers).
When the Dragon is attacked, it can attack back. With 40% probability, it will cause 1 damage to every villager in the Cave, and with 20% probability, it will eat one random villager in the Cave. Dead villagers are removed from the game. If all the villagers die, you lose the game.
The Dragon starts with 50 HP and you win the game when it dies. If you don't kill the Dragon within 30 steps, you lose the game.
\end{lstlisting}

\subsubsection{Architecture specification}
\label{sec:approach:architecture}

The architecture specification serves to formally define the components and ensembles in the system, the architecture of the system (permitted assignments of components into ensembles -- ensemble resolution), and to express the desired adaptation strategy in ensemble resolution.
It is written in ADSL that mixes formal constructs and natural language formulation snippets, as highlighted by color coding in \autoref{lst:dsl-dragon}: \blue{BLUE} denotes keywords, BLACK identifiers and expressions, and \green{GREEN} natural language snippets.

\begin{lstlisting}[style=dsl,label=lst:dsl-dragon,caption= Dragon Hunt architecture specification in ADSL (a fragment).]
# component definitions
component Dragon
  attribute hp
    name "Health"

component Villager
  name |\color{black}name|
  attribute role
    name "role"
    description "'Farmer' or 'Warrior'"
  ...

# ensemble definitions
ensemble Farm
  name "farm"
  description "Stay in the Village and work on the farm"
ensemble GoToCave
  name "cave"
  description: "Go to the Cave"
...


# observable-only environment state
beyond-control Dragon dragon "The Dragon"
...

# ensemble resolution
periodically assign Villager[] "Villagers in the Village"
  if location == 'Village'
into ensembles Farm, GoToCave, ...
as assign_in_village

periodically assign Villager[] "Villagers in the Cave"
   if location == 'Cave'
into ensembles ...
as assign_in_cave

strategy: "All Warriors should go to the Cave, and then attack the Dragon. All Farmers should stay in Village and farm or spawn new villagers (both Farmers and Warriors are necessary)."

# adaptation manager interface
am_interface SmartAdaptation(dragon.DragonHuntAdaptation)
\end{lstlisting}

\subsubsection{Prompt content}
\label{sec:approach:prompt}

An example of a generated prompt is shown in \autoref{lst:prompt-dragon}, where the \violet{VIOLET} text indicates the information obtained from the ADSL architecture specification and the domain description, and the BLACK text is part of the template used to generate the prompt.

The prompt starts with the domain description that includes the overall goal of the system. 

Next, the LLM task is articulated: Design and implement an adaptation manager that assigns components to groups\footnote{In the prompts, we use the term ``group'' instead of ``ensemble'', assuming this is less ambiguous for the LLM.}. The generic constraint that 
assigned to exactly one group'' is explicitly mentioned.

For the generated AM to be practically useful by being plugged into the system, technical requirements are mentioned, such as implementing the AM in Python and deriving it from a given abstract class. The abstract class is automatically generated from the ensemble assignment definitions (one method per assignment) specified in the ADSL.

Then, more details on each ensemble assignment are given. The available ensembles with their descriptions are listed to provide the LLM with the possible tasks for the components.

We also instruct the LLM on what attributes of the components are relevant to the adaptation and how to access them (via their identifier). As this prompt is generated at design time, it is not possible to include the attribute values for the components, so we rather instruct the LLM on how these values can be obtained in the generated AM.

In a similar vein, beyond-control components and their attributes are listed after the ensemble assignments.

Lastly, the desired strategy that the generated AM follows
and a list of functional constraints (termed ``requirements'' in the prompt)
is included in the prompt.
Functional constraints are expressed in natural language; their formal specification in FCL is then used for verification (\autoref{sec:approach:verifier}).

\begin{lstlisting}[style=prompt,label=lst:prompt-dragon,caption=Example of a generated prompt.]
|\violet{You play a computer game. ... If you don't kill the Dragon within 30 steps, you lose the game.}|

Suggest an adaptation manager. The goal is to assign the components into groups. Note that each component must be assigned to exactly one group. If a component is supposed to remain in the same group (continue performing the same action), it must always be explicitly re-assigned to that group.

The adaptation manager must be written in Python and it must be a class named `|\violet{SmartAdaptation}|` derived from this base class (can be imported from `|\violet{dragon}|`):
```
class |\violet{DragonHuntAdaptation}|(abc.ABC):
    @abc.abstractmethod
    def |\violet{assign\_in\_village}|(self, components, environment, group_ids, step):
        pass
    @abc.abstractmethod
    def |\violet{assign\_in\_cave}|(self, components, environment, group_ids, step):
        pass
```
To perform the group assignments, use the `environment.assign_group(component, group_id)` method. The `group_id` must be exactly as listed below.
---
In `|\violet{assign\_in\_village}|`, your goal is to divide the |\violet{Villagers in the Village}| (`components`) into the following groups:
- A group named "|\violet{farm}|": |\violet{Stay in the Village and work on the farm}|
- A group named "|\violet{cave}|": |\violet{Go to the Cave}|
- A group named "|\violet{spawn farmer}|": |\violet{For every two villagers assigned to this group and 10 wheat, a new Farmer is spawned.}|
- A group named "|\violet{spawn warrior}|": |\violet{For every two villagers assigned to this group and 12 wheat, a new Warrior is spawned.}|

The `group_ids` argument is a list of all valid group names.

For each component, the following attributes are available (note that the attributes are read-only and they do not update when a component is assigned to a group):
- `|\violet{role}|`: |\violet{Role}| (|\violet{'Farmer' or 'Warrior'}|)
- `|\violet{hp}|`: |\violet{Health}|
---
In `|\violet{assign\_in\_cave}|`, your goal is to divide the |\violet{Villagers in the Cave}| (`components`) into the following groups:
- A group named "|\violet{attack}|": |\violet{Attack the Dragon}|
- A group named "|\violet{cave}|": |\violet{Stay in the Cave}|
- A group named "|\violet{village}|": |\violet{Go to the Village}|

The `group_ids` argument is a list of all valid group names.

For each component, the following attributes are available (note that the attributes are read-only and they do not update when a component is assigned to a group):
- `|\violet{role}|`: |\violet{Role}| (|\violet{'Farmer' or 'Warrior'}|)
- `|\violet{hp}|`: |\violet{Health}|
---
Further, you can access the following beyond-control components, which are only observable and cannot be assigned to groups.

|\violet{The Dragon}| (accessible via `environment.|\violet{dragon}|`) with the following attributes (note that the attributes are read-only and they do not update when a component is assigned to a group):
- `|\violet{hp}|`: |\violet{Health}|
|\violet{The Farm}| (accessible via `environment.|\violet{farm}|`) with the following attributes (note that the attributes are read-only and they do not update when a component is assigned to a group):
- `|\violet{wheat}|`: |\violet{Current wheat amount}|
---
|\violet{All Warriors should go to the Cave, and then attack the Dragon. All Farmers should stay in the Village and farm or spawn new villagers (both Farmers and Warriors are necessary).}|

The adaptation strategy must adhere to the following functional requirements:
- |\violet{The Dragon should be attacked at least once in the first 15 steps of the game.}|
- |\violet{All farmers should stay in the Village.}|

Think step by step. First, reason about the task and analyze the problem. Then, describe the adaptation manager. After that, write the Python code for the adaptation manager.
\end{lstlisting}

\subsection{Verification of the AM in system runs}
\label{sec:approach:verifier}

Verification of the AM is performed by running a series of automated tests, each targeting the violation of a specific type of constraint. 
In our approach, a test consists of running the system (the adaptation loop) and checking that the trace corresponding to the system run complies with the constraints. In the running test, the Constraint Verifier (CV) checks each of the proposed architecture updates produced by the AM.
If the CV detects a constraint violation, it records it in the constraint violation report. After all tests are completed, if there are any constraints violated, the feedback loop is activated, and the report is sent to the LLM.

To achieve a better trace coverage in the verification (i.e., cover the behavior of AM in more possible system states), the tests are parameterized by \textbf{initial states} (\autoref{lst:initialsettings}). Each executed test is thus a new run of the system starting from a given initial state (e.g., the Dragon Hunt game is played from ``No Warriors'' state (\autoref{lst:initialsettings})).

\begin{lstlisting}[style=dsl,label=lst:initialsettings,caption=Initial states for verification in Dragon Hunt (fragment)]
initial state "Farmers and Warriors"
  random_seed: 42|\quad|# to guarantee repeatable experiments
  farmer_count: 2
  warrior_count: 1
initial state "No Warriors"
  random_seed: 123
  farmer_count: 2
  warrior_count: 0
\end{lstlisting}

The \textbf{generic constraints} (domain-independent)
ensure that the produced AM and its current output (architecture update/ensemble resolution) are sound in accordance with the specification expressed in the prompt. This includes testing that
\begin{itemize}
    \item the LLM response includes a block of Python code which is executable and can be imported into the system (e.g., the AM class is named correctly and derived from the abstract base class specified in the prompt),
    \item no exceptions have been thrown at runtime,
    \item only the appropriate ensembles mentioned in the architecture specification are used, e.g., in \lstinline{assign_in_cave}, the appropriate ensembles are \lstinline{Attack}, \lstinline{StayInCave}, and \lstinline{GoToVillage}),
    \item each component is assigned to exactly one ensemble.
\end{itemize}

Furthermore, \textbf{functional constraints} (domain-specific)
are checked to ensure that the assignment of components into ensembles corresponds to the strategy drawn by the architect in ADSL. Although the strategy (specified in natural language) is given to the LLM in the initial prompt, there is no guarantee that the LLM will follow it in the generated AM. The functional constraints thus allow one to formalize the requirements on the behavior of the AM, and check, in an automatized way,  whether they are met. 

The functional constraints are specified in the Functional Constraints DSL (FCDSL), which combines formally expressing the constraints in the temporal logic FCL and articulating their essence in natural language, as illustrated in~\autoref{lst:constraints-dragon}.
The former serves as input to the CV, the latter as the base for forming an entry in the constraint violation report (feedback to the LLM). Details on FCL and examples of functional constraints in FCL are given in \autoref{sec:constraints}.

\begin{lstlisting}[style=dsl,label=lst:constraints-dragon,caption=Functional constraints specification for the Dragon Hunt (a fragment in FCDSL).]
constraint "The Dragon should be attacked at least once in the first 15 steps of the game."
    |$\lozenge^1_{15} \vert\mathit{Attack}\vert \ge 1$|

constraint "All farmers should stay in the village."
    |$\mathit{Farmers} = \{ v \in \mathit{Villagers} \mid \mathit{v.role} = \text{`Farmer'} \}$|
    |$\forall f \in \mathit{Farmers}: \lozenge^{\MAX}_{\MAX}\ \mathit{f.location} = \text{`Village'}$|


\end{lstlisting}

\section{Functional Constraints Logic (FCL)}
\label{sec:constraints}
As already stated in the previous sections, the system architect can specify the desired behavior of the adaptation manager produced by vibe coding.
The strategy is written in natural language, and it is presented to the LLM when generating the AM.

Although current LLMs are trained to follow instructions, there is no guarantee that the LLM will correctly interpret the strategy and implement the AM according to it. To this end, our approach includes functional constraints as a way to formalize the behavior of the AM. The constraints allow us to formally specify expectations on the behavior via logical formulas that should hold for the system. The syntax and semantics of \textit{Functional constraints logic} (FCL) are described in \autoref{sec:constraints:logic} and examples are given in \autoref{sec:constraints:examples}.

The functional constraints are used for runtime verification of the generated AM and providing feedback to vibe coding, as already described in \autoref{sec:approach:verifier}. To ensure that the generated AM follows the desired strategy, the functional constraints have to be aligned with the strategy (both are defined by the architect).

Apart from the runtime verification, FCL could also be used to guide future adaptations of the system, which we discuss in \autoref{sec:constraints:adaptation}.

\subsection{FCL syntax and semantics}
\label{sec:constraints:logic}

FCL is based on first-order temporal logic and is inspired by TLA+~\cite{lamport2002specifying} and \cite{horizon}. We assume that the system runs in discrete time steps and allow for both finite and infinite traces.
The core of FCL is the temporal operator (called \textit{within}) $\lozenge^n_t \varphi$, expressing that the formula $\varphi$ holds at least $n$ times within the next $t > 0$ steps or at least $n$ times in the previous $-t$ steps for $t<0$. As to the relationship of $t$ (\textit{time window}) and $n$,  we define the semantics of $\lozenge^n_t \varphi$ for $n>\left|t\right|$ as $\mathit{false}$.

When considering finite runs of the system (finite traces), a special value $\MAX$ is used to denote the number of time steps from the current step to the end of the trace. Likewise, $\BEG$ denotes the number of time steps from the beginning of the trace to the current step. Thus, $\BEG$ and $\MAX$ serve as endcounts, since their value is determined by the current time step.

Apart from the temporal operator \textit{within}, the standard logical operators ($\wedge$, $\vee$, $\implies$, $\neg$), constants $\mathit{true}$ and $\mathit{false}$, quantifiers ($\forall$, $\exists$), and sets with standard operations (similar to TLA+)  are employed in FCL.

Thus, considering a system composed of components and ensembles, its state is represented by sets updated in each adaptation step (FCL time step).  For example, all instances of the \lstinline{Villager} component are available as members of the $\mathit{Villagers}$ set. The set corresponding to an ensemble is updated in each adaptation step to contain the components assigned to the ensemble, e.g., $\mathit{Farm}$ is the set of all components assigned to the \lstinline{Farm} ensemble in the current adaptation step. The attributes of the components are accessed via the dot notation (similarly to Python).

Working with finite traces requires handling the semantics of $t$ and $n$ when colliding with the window boundaries: 
$\lozenge^n_t \varphi$ for $t > \MAX \wedge n \le \MAX$ becomes semantically 
$\lozenge^n_{MAX} \varphi$ (and similarly $\lozenge^n_t \varphi$ for $t < 0$,  $\left|t\right| > \BEG$ and $n \le \BEG$ becomes semantically
$\lozenge^n_{BEG}\varphi$).

As an aside, using the $\lozenge^n_t$ operator and $\INF$ denoting infinity, the common LTL operators can be expressed as follows:
\begin{itemize}[itemsep=4pt]
    \item \textit{next} ($\varphi$ holds in the next time step): $\bigcirc \varphi \triangleq \lozenge^1_1 \varphi$,
    \item \textit{future} ($\varphi$ holds at least once in the future): $\lozenge \varphi \triangleq \lozenge^1_\INF \varphi$,
    \item \textit{globally} ($\varphi$ holds in each future step): $\square \varphi \triangleq \lozenge^\INF_\INF\varphi$
\end{itemize}

By general, the semantics of \textit{within} applied to the LTL \textit{until} operator and the nesting of \textit{within} can be inductively defined by the concept of \textit{horizon} (basically the maximal time window for which a composed formula has to hold) as in~\cite{horizon}. Given the main focus of this text and its space constraints, we do not elaborate on such formal definitions further.

\subsection{Examples of functional constraints for Dragon Hunt in FCL}
\label{sec:constraints:examples}

In the examples below, the following sets represent the system's components: $\mathit{Dragons}$, $\mathit{Villagers}$; and ensembles: $\mathit{Farm}$, $\mathit{GoToCave}$, $\mathit{SpawnFarmer}$, $\mathit{SpawnWarrior}$, $\mathit{Attack}$, $\mathit{StayInCave}$, $\mathit{GoToVillage}$.

\subsubsection{The game is won}

The game is won when the Dragon is killed, i.e., when its health points reach zero. The set $\mathit{Dragons}$ contains all components of type \lstinline{Dragon} (only one in this case). We use the \textit{within} operator to express that the Dragon's health reaches zero at some point of the game (technically, at least once).
\[\forall d \in \mathit{Dragons}: \lozenge^1_{\MAX}\ d.hp \le 0\]

\subsubsection{The Dragon is attacked at least once within the first 15 steps of the game}

To express that the Dragon is attacked at least once during the first 15 steps of the run of the system, we check that the \lstinline{Attack} ensemble has at least one member (its cardinality is $\ge 1$).
\[\lozenge^1_{15} |\mathit{Attack}| \ge 1\]

\subsubsection{All farmers should stay in the Village}

Our strategy desires that all farmers stay in the village. First, we define a set of all farmers by selecting some of the \lstinline{Villager} components based on their role.
\[\mathit{Farmers} = \{ v \in \mathit{Villagers} \mid \mathit{v.role} = \text{``Farmer''} \}\]

Then, we use $\lozenge^{\MAX}_{\MAX}$ to express that it should always hold (in every time step) that the farmer's location is the village.
\[\forall f \in \mathit{Farmers}: \lozenge^{\MAX}_{\MAX}\ \mathit{f.location} = \text{``Village''}\]

\subsubsection{All warriors should go to the Cave}

Each warrior should be in the \lstinline{GoToCave} ensemble (i.e., go from the Village to the Cave) at least once.
\begin{align*}
&\mathit{Warriors} = \{ v \in \mathit{Villagers} \mid \mathit{v.role} = \text{``Warrior''} \}\\
&\forall w \in \mathit{Warriors}: \lozenge^{1}_\MAX\ w \in \mathit{GoToCave}
\end{align*}

\subsubsection{Spawn at least a few new warriors}
\label{sec:constraints:examples:spawn_warriors}

To effectively kill the dragon, we want at least three warriors to be spawned during the run of the system. The rules of the game define that two villagers are necessary to spawn a new one. The \lstinline{SpawnWarrior} ensemble should therefore have at least two members at least three times.
\[\lozenge^3_{\MAX} |\mathit{SpawnWarrior}| \ge 2\]

\subsubsection{Spawn at least a few new farmers}

(similar to above)
\[\lozenge^3_{\MAX} |\mathit{SpawnFarmer}| \ge 2\]

\subsubsection{After a villager gets to the cave, it should attack the dragon}

An implication can be used to express a sequence of actions. If a component is a member of the \lstinline{GoToCave} ensemble, it moves from the Village to the Cave. We want such components to attack the Dragon at some point in the future (i.e., at least once be a member of the \lstinline{Attack} ensemble).
\[\forall v \in \mathit{Villagers}: v \in \mathit{GoToCave} \implies \lozenge^1_\MAX\ v \in \mathit{Attack}\]
However, it can happen that a villager moves to the Cave in the last step of the trace, which leaves no more steps for them to attack the Dragon, leading to constraint violation. To prevent this, we update the implication to not enforce it in the last step (i.e., when $MAX = 0$).
\[\left(v \in \mathit{GoToCave} \wedge \MAX>0 \right) \implies \lozenge^1_\MAX\ v \in \mathit{Attack}\]

\subsubsection{Warriors should be mostly in the cave}

We also want the warriors to spend most of the time in the Cave, but the fact that they are spawned in the Village and first need to go to the cave prevents us from using the ``always'' operator ($\lozenge^{\MAX}_{\MAX}$). We thus formulate the constraint as follows: at least 80\% of the time (i.e., $0.8$ times the number of remaining steps), at least half of the warriors are in the Cave.
\begin{align*}
&\mathit{Warriors} = \{ v \in \mathit{Villagers} \mid \mathit{v.role} = \text{``Warrior''} \}\\
&\mathit{InCave} = \{ v \in \mathit{Villagers} \mid \mathit{v.location} = \text{``Cave''} \}\\
&\lozenge^{0.8 \cdot \MAX}_\MAX |\mathit{Warriors} \cap \mathit{InCave}| \ge 0.5 \cdot |\mathit{Warriors}|
\end{align*}

\subsubsection{Spawn a new warrior every 10 steps}\footnote{In the end, this constraint turned out to be too strong and was not imposed in the Dragon Hunt later on.}

To express this constraint, we reformulate it as follows. If no warrior was spawned in the previous 10 steps, spawn a warrior. We use $t=-10$ and $n=10$ to express that the \lstinline{SpawnWarrior} ensemble had less than 2 members in each of the previous 10 steps.
\[\lozenge^{10}_{-10} |\mathit{SpawnWarrior}| < 2 \implies |\mathit{SpawnWarrior}| \ge 2\]

\section{Evaluation}
\label{sec:evaluation}
\subsection{Tools and experiment setup}

The main aim of our evaluation was to show that vibe coding with feedback verification is a viable approach. To this end, we implemented a framework that realizes all the conceptual components depicted in \autoref{fig:approach} and allows us to control the verification process described in \autoref{sec:approach:verifier}.  Thus, it constructs the prompts (\autoref{sec:approach:prompt}), provides input to the LLM through an API\footnote{\href{https://openai.com/api/}{\texttt{openai.com/api}}}, runs the system adaptation loop with constraint verification, and activates the feedback loop by presenting feedback to the LLM with a constraint violation report. 

The main criterion observed in each experiment
is the number of feedback loop iterations needed to obtain a \textit{valid} AM (one that does not violate any of the constraints imposed). 
Since it is not guaranteed that the LLM will ever produce a valid AM, we also abort an experiment after 10 unsuccessful iterations.

For the evaluation of the Dragon Hunt example, the framework was used as follows:
First, in a series of initial experiments, we tuned up the functional constraints (their final form is in \autoref{sec:constraints:examples}) so that if the   
AM loses the game, there should be a constraint violation explaining why this has happened (more on the tuning process is in the replication package\replication).

Second, we run a series of experiments aiming at comparing different forms of feedback to LLM, i.e., the content of the constraint violation report:
\begin{enumerate}
    \item generic and functional constraint violations,
    \item generic constraint violations only,
    \item baseline: instead of constraint violations, the feedback contained system domain metrics (for Dragon Hunt the win rate and the number of steps to win the game).
\end{enumerate}
The variants differ only in the feedback presented to LLM; for verification (i.e., determining if AM is valid), all generic and functional constraints are always used.

Moreover, we used two versions of the initial prompt:
\begin{itemize}
    \item[a)] functional constraints already in the initial prompt (as shown in \autoref{sec:approach:prompt}),
    \item[b)] no functional constraints mentioned in the initial prompt,
\end{itemize}
Each form of feedback was combined with each initial prompt, leading to six experiment variants in total.

Here, since the Dragon Hunt example is relatively simple, we opted for the ``GPT-5 nano'' LLM (version \lstinline{gpt-5-nano-2025-08-07}), supposedly a less capable LLM likely needing more than one iteration to produce a valid AM.

Finally, to take into account the stochasticity of LLM responses, we repeated each experiment 10 times.

The distribution of the results collected from these 10 experiments for the 6 experiment variants is shown in~\autoref{fig:results}; the results are discussed in \autoref{sec:evaluation:results}.
The implementation of the framework, as well as the experiment setup and the raw results are available in the replication package\replication.

\subsection{FCL implementation}

As part of the Constraint Verifier, we implemented a subset of FCL that was sufficient to express all the functional constraints specified in \autoref{sec:constraints:examples}. It is a simplified FCL version allowing only one implication in a constraint and prohibiting nested \textit{within} operators (recall that the meaning of the \textit{within} temporal operator $\lozenge^n_t\varphi$ is that the formula $\varphi$ must be true at least $n$ times within the next $t$ steps). Furthermore, in a constraint, the \textit{within} operator with $t>0$ can be either the top-level syntactical construct or in the consequent of a syntactically top-level implication. For $t<0$, the \textit{within} operator can be in the antecedent of such an implication.

The evaluation of FCL constraints is done online while the system runs. As for the \textit{within} operator $\lozenge^n_t\varphi$ with $t<0$, a history of the previous $-t$ steps is recorded to evaluate it in the current step. For $t>0$, the \textit{within} operator cannot be evaluated immediately, since the needed future values are not yet known. This is implemented by creating a \textit{temporal obligation} anytime the \textit{within} operator is to be evaluated (i.e., typically at the beginning of the run if it is syntactically a top-level construct, or when the antecedent of an implication is true). 
In each step, $\varphi$ is evaluated and the result is recorded. The temporal obligation keeps track of the number of time steps in which $\varphi$ was true. At the end of the time window (after $t$ steps) or if it is already possible to determine the result, e.g. $\varphi$ was already true $n$ times or $\lozenge^n_t\varphi$ cannot be true (i.e., $\varphi$ is false more than $t-n$ times), the evaluation of the obligation ends.

\subsection{Results}
\label{sec:evaluation:results}

\begin{figure}[tbh]
	\includegraphics[width=\linewidth]{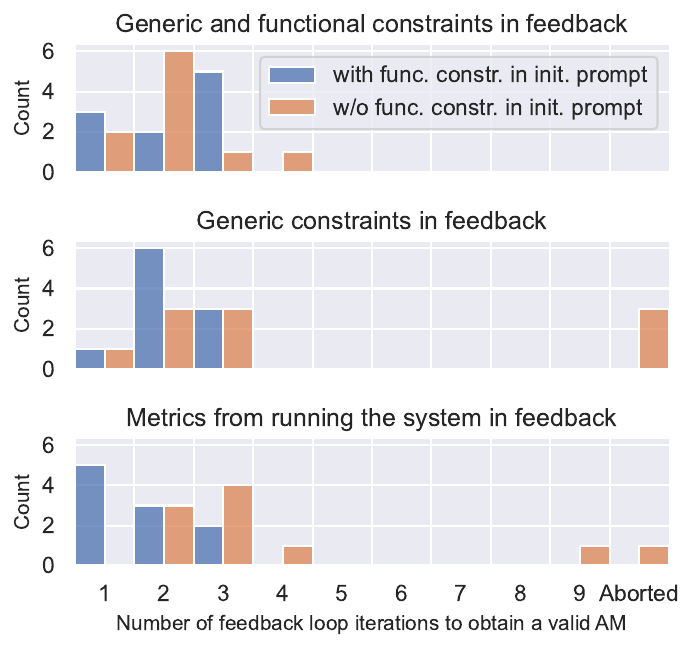}
    \caption{Distributions of the number of feedback-loop iterations needed to obtain a valid AM in Dragon Hunt example. Bars show counts over 10 experiments for each variant.}
	\label{fig:results}
\end{figure}

Figure~\ref{fig:results} shows a series of histograms of the number of iterations of the vibe-coding feedback loop necessary to produce a valid AM for three variants of feedback to LLM (rows) and two variants of the initial prompt (colors).
The more the histogram is concentrated toward the left, the better.

The results indicate that combining vibe coding with constraint verification is an effective form of generating an AM for the system. In most of the experiment runs, the LLM managed to produce a correctly working AM behaving according to the desired strategy (a valid AM) in just a few iterations of the feedback loop. However, for this approach to be useful, it is necessary to align functional constraints with the goals of the system, as we discuss more in \autoref{sec:evaluation:constraints-alignment}. In this case, we managed to design the constraints in such a way that all generated AMs meeting the constraints won the game.

Regarding different forms of feedback, in the Dragon Hunt example, there does not seem to be much difference when the functional constraints are listed in the initial prompt. When the list is removed, reporting violations of both generic and functional constraints performs the best, as for the other two methods, there were at least a few experiment runs that did not produce a valid AM (those were aborted after 10 iterations of the feedback loop). This was expected as the constraints were also used to determine the validity of the generated AM. However, this also indicates that ``insufficient'' feedback might cause the feedback loop to get stuck in a dead end (\autoref{sec:evaluation:dead-end}).

\subsection{Validation on a second example: Smart Farm}

Apart from running the experiments with the Dragoon Hunt example, we also evaluated our approach for a second example: the  Smart Farm. Its scenario comes from
one of our recently finished ECSEL JU project AFarCloud\footnote{\url{http://www.afarcloud.eu/}}
and focuses on protecting crop fields on a farm from feeding birds by groups of autonomous drones.
As birds fly from one field to another, the drones need to be coordinated and redistributed among the fields to ensure protection of the most-threatened fields. If a field is not adequately protected, the birds cause damage to the field. The overall goal is to minimize the damage rate (in \% of the field area).

This example is more complex than the Dragon Hunt, as it is necessary to take into account the positions of the drones when redistributing them (i.e., the villagers with the same role were interchangeable, but the drones are not). Also, there are several instances of a protecting ensemble (one for each field) compared to only singleton ensembles in the Dragon Hunt.

As GPT-5 nano often failed to produce a valid AM in this harder example, we also experimented with a more capable LLM ``GPT-5 mini'' (version \lstinline{gpt-5-mini-2025-08-07}).

Similarly to the Dragon Hunt example, we designed a set of functional constraints 
(and refined them after initial experiments) for the AM to follow in order to ensure adequate protection of the fields. We used 6 constraints and they are listed in the replication package\replication.

\begin{figure}[tbh]
	\includegraphics[width=\linewidth]{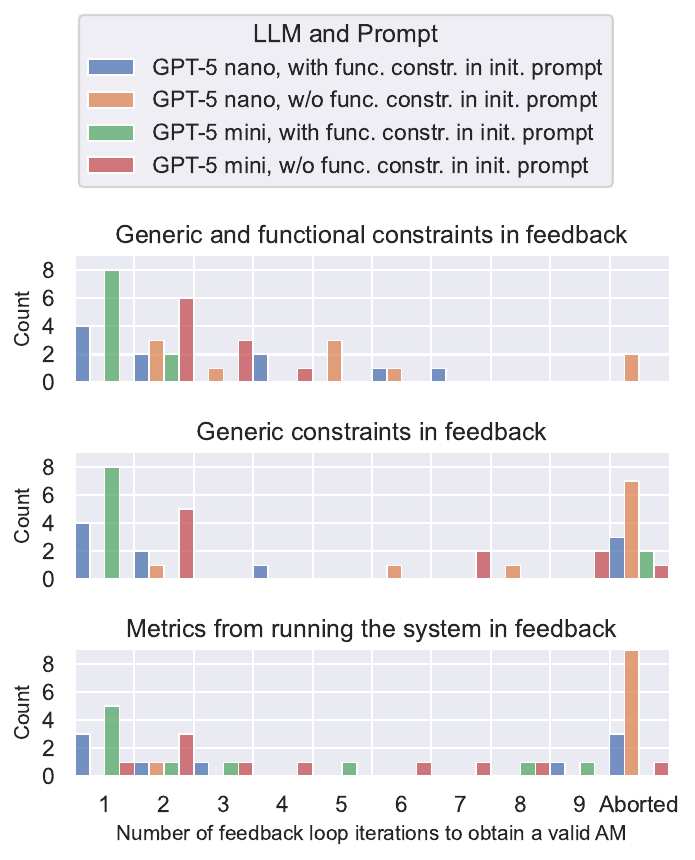}
    \caption{Distributions of the number of feedback-loop iterations needed to obtain a valid AM in Smart Farm example. Bars show counts over 10 independent experiments for each variant.}
	\label{fig:results_farm}
\end{figure}

The results, summarized in \autoref{fig:results_farm}, show a pattern similar to that in the Dragon Hunt example. Providing generic and functional constraint violations as feedback to the LLM mostly leads to a valid AM in a few iterations. The other forms of feedback perform worse. Without functional constraints listed in the initial prompt or given as feedback to the LLM, the GPT-5 nano frequently fails to generate a valid AM.

Regarding the efficiency of the generated AMs in protecting the fields, they perform almost as well as our previously hand-written AM: The median of damage rate for valid generated AMs was $6.9$\%, while for the hand-written AM it was $7.0$\%.

\subsection{Discussion}

In this section, we discuss several lessons learned from setting up the vibe coding feedback loop and expressing constraints in FCL.

\subsubsection{Results summary}

When the functional constraints are listed in the initial prompt, the LLMs seem to be very capable in producing a valid AM. However, especially with less capable LLMs, it is still worth verifying the AM and checking whether the requirements are met.

Regarding the different forms of feedback to the LLM we experimented with, giving the LLM detailed feedback (violations of generic and functional constraints) performs the best as the other two forms of feedback did not give the LLM enough information on what should be improved to produce a valid AM.

\subsubsection{Functional constraints as precise as possible}
\label{sec:evaluation:constraints-alignment}

Several times in our initial experiments, the LLM designed an AM that met all the constraints, but did not win the game. By further inspecting the behavior of the AM, we noticed that the game had been lost only by a small margin (the dragon survived with only a few health points). In particular, the AM did not spawn warriors often enough to win the game; however, it managed to spawn 3 of them during the run, thus fulfilling the constraint.

To mitigate this issue, we updated the set of functional constraints and initial states, adding ``The dragon is attacked at least once within the first 15 steps of the game'' constraint and ``No Warriors'' initial state. After that, the issue was solved.

The takeaway here is that 
it might be necessary in the feedback loop to re-formulate the constraints in alignment with the desired strategy and update appropriate run path coverage (human involvement in the feedback loop  might be necessary as also reported in~\cite{ge2025surveyvibecodinglarge}).

\subsubsection{Feedback loop reaching a dead end}
\label{sec:evaluation:dead-end}

\begin{figure}[tbh]
	\includegraphics[width=\linewidth]{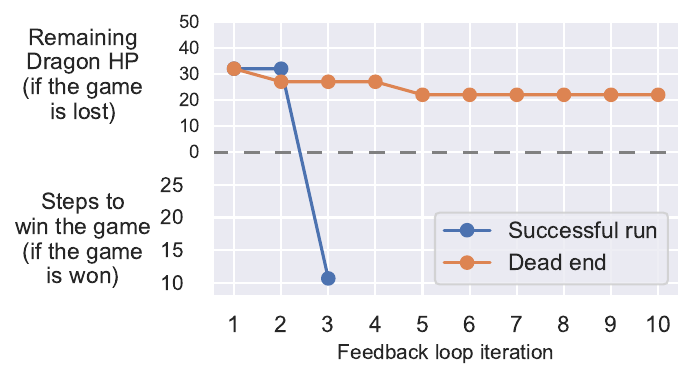}
	\caption{Two runs of an experiment -- one successful (valid AM after 3 iterations) and one reaching a dead end and not improving anymore.}
	\label{fig:dead-end}
\end{figure}

We encountered several times that the vibe-coding feedback loop reached a dead end, so that the AM did not improve anymore. This often happened when there was insufficient feedback. For instance, when the feedback is in the form of metrics from running the system, the only information given to the LLM is that it lost (``win rate is 0\%''). As this feedback repeated multiple times in a row, the LLM tended to produce a very similar AM. It might have been slightly modified in terms of exact implementation, but it actually behaved the same. This is illustrated in \autoref{fig:dead-end}.

The key takeaway here is that the feedback to the LLM needs to be precise enough. If we only said ``win rate is 0\%'', the LLM could not identify the reason for the loss and, consequently, how to find a solution to such AM malfunction.

In such situations, we found it might be better to break the feedback loop and start over, since with a fresh start, the LLM is not confused by the conversation history with the previous unsuccessful attempts into producing a similar code.

\subsubsection{Importance of fine-grained constraints thanks to FCL} 
As mentioned in the discussion above, specific and detailed-enough feedback in the feedback loop is crucial for a successful vibe coding employment. In this respect, FCL appeared to be very useful in allowing expression of the desired properties/constraints at the level of multiple individual adaptation steps in a time window. This is in contrast with the expressive power of constraint granularity in, e.g. classical temporal logics such as LTL (considering only next, future, globally, until) and in TLA+ modeling the advance of execution at the level of just neighboring states (the current and the ``primed'' (new) state). 

It should be emphasized that the runtime implementation of FCL constraints based on the concept of temporal obligation turned out to be relatively easy. As an aside, the idea of endcounts ($\MAX$ and $\BEG$) necessary to model the relationship of multiple steps in finite traces was also implemented in this way.

\subsubsection{AM could cooperate with FCL interpreter}
\label{sec:constraints:adaptation}

Another potential use for FCL, apart from constraint verification, is to prioritize AM adaptation decisions based on the access of the AM to the temporal obligations in the FCL interpreter of Constraint Verifier. 

Take, for example, the constraint $\lozenge^1_{\MAX} |\mathit{Attack}| \ge 1$. Consider the very last step of the run, and this constraint still not being satisfied -- this means that the \lstinline{Attack} ensemble must not be empty in this step, since otherwise the constraint would be violated. Assuming AM had access to the temporal obligation associated with the constraint, it could assign some components to the \lstinline{Attack} ensemble to make the constraint satisfied in this last-minute opportunity.  
Furthermore, if the AM was sufficiently sophisticated, it could have anticipated this and sent some more villagers to the Cave in the previous steps since only the villagers in the Cave can be assigned to the \lstinline{Attack} ensemble. 

Fine-grained adaptation rules such as these can be dynamically inferred from the FCL temporal obligations to guide the AM by narrowing potential adaptation options towards efficiently meeting the statically specified strategy in ADSL. For now, this is left as a potential direction for future work.

\subsection{Limitations and Threats to Validity}
\label{sec:threats}

We structure threats to validity according to~\cite{Runeson2008}:

\subsubsection{Construct validity}

Although our results indicate that the approach presented in this paper is viable for the two simple examples with which we experimented, we did not perform any experiments with real-world CAS. In theory, the behavior of a more complex system could be described with more elaborate FCL constraints, but designing such constraints might be demanding, as it is necessary to describe the system in enough detail and align the constraints with the goals of the system.

Another potential threat arises from our definition of a valid AM in the experiment. We used the functional constraints to determine whether the generated AM follows the desired strategy we specified in the prompt in natural language. Although we did our best to formulate the constraints according to the strategy and validated them using initial experiments, we did not inspect the LLM-generated code in detail. It is theoretically possible that the LLM generated a code that does not behave exactly according to the desired strategy, but still passes the constraints. Nevertheless, we also observed the results of running the examples with the generated AMs and found out that all the AMs approved as valid by the runtime verification performed well in terms of metrics of the system (e.g., managed to win the game by killing the Dragon).

Lastly, the experiments in this paper included only a small subset of available LLMs, so that other LLMs might perform differently. However, the main contribution of this paper is the approach that combines vibe coding and FCL-based runtime verification, and that does not depend on a specific LLM (the LLM used can be simply swapped by a slight modification of the API calls). 

\subsubsection{Internal validity}

The main threat in this area is the stochasticity of both the LLM responses and our use case implementation. We tried to mitigate this by repeating the experiments several times and by observing the trends among the aggregated results.

\subsubsection{External validity}

We tried to design the approach to be general enough and applicable to other use cases by working on top of the abstractions of the DEECo model, which has already proven to be suitable for modeling CAS, and our novel temporal logic FCL. The approach uses a DSL to specify the domain-specific details of the use case, and the LLMs prompts are automatically generated from them, making it relatively easy to apply them in other domains.
In line with this objective, we used two different use case examples for the evaluation of our approach.

\subsubsection{Reliability}

Reliability threats arise from the fact that LLMs can be sensitive to exact formulations of instructions in the prompts. Thus, altering the wording of the instructions and the way the prompts are constructed might change the results.

\section{Related Work}
\label{sec:related}
Several related works survey \textbf{vibe coding} techniques~\cite{ge2025surveyvibecodinglarge} and list different forms of feedback to the LLM.
Common options are to perform static analysis of the code and run unit tests.
Liang et al.~\cite{Liang2025} iteratively refine the generated code while also generating the tests with an LLM. Nunez et al.~\cite{nunez2024autosafecodermultiagentframeworksecuring} suggest a multi-agent framework, where three LLM-based agents are employed: one for generating the code, one for static analysis, and one for fuzzy testing.
Ravi et al.~\cite{inproceedings} suggest five automatic feedback loops addressing compilation errors, static analysis, hand-written unit tests, and generated unit tests with mutation analysis. Kavian et al.~\cite{Kavian2024} combine LLMs with static code analysis engines to improve the security of the generated code. In addition, checking for architectural smells by LLMs proposed by Pandini et al.~\cite{architsmells} can be used to improve the code. 

Similarly to our approach, several other works propose some form of \textbf{runtime verification} in vibe coding. Councilman et al.~\cite{councilman2025formalverificationllmgeneratedcode} propose a high-level formal query language, which expresses the user’s intent in structured natural language, and a verifier. In contrast, we use temporal logic for the user’s intent.
Zhang et al.~\cite{zhang2025rvllmllmruntimeverification} design a general specification language based on predicate logic that enables automatic checking whether LLM-generated responses adhere to domain-specific properties. However, their focus is not on vibe-coding, they verify the LLMs responses in natural language. Due to that, their proposed language does not need to consider the temporal dimension, which we do need in our approach for describing CAS.

Several works employ \textbf{logic} for expressing and checking properties of CAS (an an overview of different approaches to specifying and modeling CAS can be found in \cite{Denicola}). Since they use logic as means to specify the behavior of such a system as a whole, they use modal logic that employs dynamic, spatial, and temporal concepts, such as dL \cite{Platzer} and SSTL \cite{SSTLlogic}. Of these, the most related to our approach is the GLoTL logic \cite{horizon,horizon2} introduced to specify the spatial and temporal properties of cooperating CAS agents at both the local and global levels. In particular, the authors introduce the concept of \textit{horizon}, setting the interval $[k,l]$ of time steps in which $\varphi$ in the temporal formula $\psi U^{[k,l]} \varphi$ has to hold a number of times. In comparison, our \textit{within} operator features, in addition to a similar time window, the option to explicitly specify the number of steps in which $\varphi$
has to hold and allows for working with finite traces thanks to the endcounts $\MAX$ and $\BEG$.


Regarding the \textbf{use of LLMs for adaptation}, several works focus on enhancing the MAPE-K loop with LLM-driven reasoning~\cite{Li2024,10628416} or revising the MAPE-K phases in the context of learning-enabled components~\cite{Casimiro2022}. Töpfer et al.~\cite{Tpfer2025} show how to employ LLM for ensemble resolution, but their focus (similarly to the previously mentioned works) is mainly on employing LLMs directly as part of the AM (the LLM is queried in every adaptation step). In contrast, our approach uses the LLM to generate the code for the AM before the system is run, and the LLM is not used during the adaptation. A hybrid approach is proposed by Adnan et al.~\cite{adnan2025leveragingllmsdynamiciot}, where the LLM is used in the context of an IoT system to select a service or implement a new one based on user-defined goals.

In general, combining ML techniques with software architecture improves the adaptability and reliability of ML-based systems~\cite{dagstuhl,9462033,Casimiro2022}. These works emphasize runtime validation and adaptation of ML components. We similarly employ runtime validation to ensure that the architecture adheres to the predefined constraints.

\section{Conclusion}
\label{sec:conclusion}

We conclude the paper by summarizing the way we answered the research questions articulated in \autoref{sec:introduction}.

\ref{RQ1}: We have shown that generating an AM via vibe coding is a viable option, provided that a sophisticated feedback loop is used to verify AM functionality. The verification is to be based on a very precise formulation of the behavior requirements/constraints. To this end, we designed FCL, a temporal logic that allows us to specify the behavior of traces with the granularity much finer than the classical LTL logic offers.

\ref{RQ2}: The key idea employed to address RQ2 presented in the paper is to integrate the checking of constraints, both generic and, in particular, functional, into the adaptation loop, so that each architectural adaptation proposed by the AM is first checked for being free of constraint violations.  

\ref{RQ3}: The idea of combining the adaptation and feedback loops has shown to be viable in the experiments with the two examples of CAS. In particular, the feedback specifying detailed violation  of fine-grained constraints articulated in FCL did not require more than a few feedback loop iterations, provided that such testing is combined with the aim of high run path coverage achieved by different initial settings of system runs.

\bibliographystyle{IEEEtran}
\bibliography{paper}

@misc{replication,  
title = {Replication package},
  year  = {2024},
  url   = {TODO},
}

@article{bures_language_2020,
	title = {A language and framework for dynamic component ensembles in smart systems},
	volume = {22},
	number = {4},
	journal = {Int. Journ. on Software Tools for Technology Transfer},
	author = {Bures, Tomas and Gerostathopoulos, Ilias and Hnetynka, Petr and Plasil, Frantisek and Krijt, Filip and Vinarek, Jiri and Kofron, Jan},
	year = {2020},
}

@INPROCEEDINGS {10628416,
author = { Donakanti, Raghav and Jain, Prakhar and Kulkarni, Shubham and Vaidhyanathan, Karthik },
booktitle = {Companion Proceedings of ICSA 2024},
title = {{ Reimagining Self-Adaptation in the Age of Large Language Models }},
year = {2024},
pages = {171-174},
publisher = {IEEE CS},
}

@article{dagstuhl,
  author =	{Lewis, Grace A. and Muccini, Henry and Ozkaya, Ipek and Vaidhyanathan, Karthik and Weiss, Roland and Zhu, Liming},
  title =	{{Software Architecture and Machine Learning (Dagstuhl Seminar 23302)}},
  pages =	{166--188},
  journal =	{Dagstuhl Reports},
  year =	{2024},
  volume =	{13},
  number =	{7},
  publisher =	{Schloss Dagstuhl -- Leibniz-Zentrum f{\"u}r Informatik},
  address =	{Dagstuhl, Germany},
}

@INPROCEEDINGS{9462033,
  author={Bureš, Tomáš},
  booktitle={2021 International Symposium on Software Engineering for Adaptive and Self-Managing Systems (SEAMS)}, 
  title={Self-Adaptation 2.0}, 
  year={2021},
  volume={},
  number={},
  pages={262-263},
}

@inbook{Casimiro2022,
  title = {Self-adaptive Machine Learning Systems: Research Challenges and Opportunities},
  booktitle = {Software Architecture},
  publisher = {Springer},
  author = {Casimiro,  Maria and Romano,  Paolo and Garlan,  David and Moreno,  Gabriel A. and Kang,  Eunsuk and Klein,  Mark},
  year = {2022},
  pages = {133–155}
}

@inproceedings{Li2024,
  title = {Exploring the Potential of Large Language Models in Self-adaptive Systems},
  booktitle = {Proc. of SEAMS 2024, Lisbon, Portugal},
  author = {Li,  Jialong and Zhang,  Mingyue and Li,  Nianyu and Weyns,  Danny and Jin,  Zhi and Tei,  Kenji},
  year = {2024},
  pages = {77–83},
}

@misc{adnan2025leveragingllmsdynamiciot,
      title={Leveraging LLMs for Dynamic IoT Systems Generation through Mixed-Initiative Interaction}, 
      author={Bassam Adnan and Sathvika Miryala and Aneesh Sambu and Karthik Vaidhyanathan and Martina De Sanctis and Romina Spalazzese},
      year={2025},
      url={https://arxiv.org/abs/2502.00689}, 
}

@article{Runeson2008,
  title = {Guidelines for conducting and reporting case study research in software engineering},
  volume = {14},
  ISSN = {1573-7616},
  url = {http://dx.doi.org/10.1007/s10664-008-9102-8},
  DOI = {10.1007/s10664-008-9102-8},
  number = {2},
  journal = {Empirical Software Engineering},
  publisher = {Springer Science and Business Media LLC},
  author = {Runeson,  Per and H\"{o}st,  Martin},
  year = {2008},
  month = dec,
  pages = {131–164}
}

@article{KOVALYOV20101908,
title = {A generic approach to proving NP-hardness of partition type problems},
journal = {Discrete Applied Mathematics},
volume = {158},
number = {17},
pages = {1908-1912},
year = {2010},
issn = {0166-218X},
doi = {https://doi.org/10.1016/j.dam.2010.08.001},
url = {https://www.sciencedirect.com/science/article/pii/S0166218X10002647},
author = {Mikhail Y. Kovalyov and Erwin Pesch}
}

@inproceedings{bures_deeco_2013,
    title = {{DEECo -- an Ensemble-Based Component System}},
    author = {Bures, Tomas and Gerostathopoulos, Ilias and Hnetynka, Petr and Keznikl, Jaroslav and Kit, Michal and Plasil, Frantisek},
    year = {2013},
    booktitle = {{Proc. of CBSE'13}},
    publisher = {ACM},
    isbn = {978-1-4503-2122-8},
    pages = {81--90},
}

@article{Tpfer2025,
  title = {On Limits of {LLMs} in Adaptation of Ensemble-Based Architectures},
  url = {http://dx.doi.org/10.2139/ssrn.5357551},
  DOI = {10.2139/ssrn.5357551},
  publisher = {Elsevier BV},
  author = {T\"{o}pfer,  Michal and Bureš,  Tomáš and Plášil,  František and Hnětynka,  Petr},
  year = {2025}
}

@book{lamport2002specifying,
author = {Lamport, Leslie},
title = {Specifying Systems: The TLA+ Language and Tools for Hardware and Software Engineers},
year = {2002},
month = {June},
publisher = {Addison-Wesley},
url = {https://www.microsoft.com/en-us/research/publication/specifying-systems-the-tla-language-and-tools-for-hardware-and-software-engineers/},
}

@article{mapek,
    author = {IBM},
    title = {An architectural blueprint for autonomic computing},
    journal = {Autonomic Computing White Paper},
    year = {2005}
}

@misc{ge2025surveyvibecodinglarge,
      title={A Survey of Vibe Coding with Large Language Models}, 
      author={Yuyao Ge and Lingrui Mei and Zenghao Duan and Tianhao Li and Yujia Zheng and Yiwei Wang and Lexin Wang and Jiayu Yao and Tianyu Liu and Yujun Cai and Baolong Bi and Fangda Guo and Jiafeng Guo and Shenghua Liu and Xueqi Cheng},
      year={2025},
      eprint={2510.12399},
      archivePrefix={arXiv},
      primaryClass={cs.AI},
      url={https://arxiv.org/abs/2510.12399}, 
}

@inproceedings{inproceedings,
author = {Ravi, Ravin and Bradshaw, Dylan and Ruberto, Stefano and Jahangirova, Gunel and Terragni, Valerio},
year = {2025},
month = {07},
pages = {},
title = {LLMLOOP: Improving LLM-Generated Code and Tests through Automated Iterative Feedback Loops},
booktitle = {Proceedings of ICSME 2025}
}

@article{Denicola,
  title = {Rigorous engineering of collective adaptive systems: special section},
  volume = {22},
  ISSN = {1433-2787},
  url = {http://dx.doi.org/10.1007/s10009-020-00565-0},
  DOI = {10.1007/s10009-020-00565-0},
  number = {4},
  journal = {International Journal on Software Tools for Technology Transfer},
  publisher = {Springer Science and Business Media LLC},
  author = {De Nicola,  Rocco and J\"{a}hnichen,  Stefan and Wirsing,  Martin},
  year = {2020},
  month = may,
  pages = {389–397}
}

@article{SSTLlogic,
DOI= {10.1007/s10009-018-0483-8},
author = {Ciancia, Vincenzo and Gilmore, Stephen and Grilletti, Gianluca and Latella, Diego and Loreti, Michele and Massink, Mieke},
title = {Spatio-temporal model checking of vehicular movement in public transport systems},
year = {2018},
issue_date = {June      2018},
publisher = {Springer-Verlag},
address = {Berlin, Heidelberg},
volume = {20},
number = {3},
issn = {1433-2779},
url = {https://doi.org/10.1007/s10009-018-0483-8},
journal = {Int. J. Softw. Tools Technol. Transf.},
month = jun,
pages = {289–311},
numpages = {23}
}

@inproceedings{horizon,
doi = {10.1007/978-3-031-16336-4_7},
author = {Loreti, Michele and Rehman, Aniqa},
title = {A Logical Framework for Reasoning About Local and Global Properties of Collective Systems},
booktitle = {Quantitative Evaluation of Systems: 19th International Conference, QEST 2022, Warsaw, Poland, September 12–16, 2022, Proceedings},
year = {2022},
isbn = {978-3-031-16335-7},
publisher = {Springer-Verlag},
address = {Berlin, Heidelberg},
url = {https://doi.org/10.1007/978-3-031-16336-4_7},
pages = {133–149},
numpages = {17},
location = {Warsaw, Poland},
}

@misc{councilman2025formalverificationllmgeneratedcode,
      title={Towards Formal Verification of LLM-Generated Code from Natural Language Prompts}, 
      author={Aaron Councilman and David Jiahao Fu and Aryan Gupta and Chengxiao Wang and David Grove and Yu-Xiong Wang and Vikram Adve},
      year={2025},
      eprint={2507.13290},
      archivePrefix={arXiv},
      primaryClass={cs.PL},
      url={https://arxiv.org/abs/2507.13290}, 
}

@misc{zhang2025rvllmllmruntimeverification,
      title={RvLLM: LLM Runtime Verification with Domain Knowledge}, 
      author={Yedi Zhang and Sun Yi Emma and Annabelle Lee Jia En and Jin Song Dong},
      year={2025},
      eprint={2505.18585},
      archivePrefix={arXiv},
      primaryClass={cs.AI},
      url={https://arxiv.org/abs/2505.18585}, 
}

@inbook{Liang2025,
  title = {RECODE: Leveraging Reliable Self-generated Tests and Fine-Grained Execution Feedback to Enhance LLM-Based Code Generation},
  ISBN = {9789819500147},
  ISSN = {1611-3349},
  url = {http://dx.doi.org/10.1007/978-981-95-0014-7_43},
  DOI = {10.1007/978-981-95-0014-7_43},
  booktitle = {Advanced Intelligent Computing Technology and Applications},
  publisher = {Springer Nature Singapore},
  author = {Liang,  Yunhao and Ying,  Ruixuan and Taniguchi,  Takuya and Gan,  Chengguang and Cui,  Zhe},
  year = {2025},
  pages = {510–521}
}

@inproceedings{Kavian2024,
  series = {EASE 2024},
  title = {LLM Security Guard for Code},
  url = {http://dx.doi.org/10.1145/3661167.3661263},
  DOI = {10.1145/3661167.3661263},
  booktitle = {Proceedings of the 28th International Conference on Evaluation and Assessment in Software Engineering},
  publisher = {ACM},
  author = {Kavian,  Arya and Pourhashem Kallehbasti,  Mohammad Mehdi and Kazemi,  Sajjad and Firouzi,  Ehsan and Ghafari,  Mohammad},
  year = {2024},
  month = jun,
  pages = {600–603},
  collection = {EASE 2024}
}

@misc{nunez2024autosafecodermultiagentframeworksecuring,
      title={AutoSafeCoder: A Multi-Agent Framework for Securing LLM Code Generation through Static Analysis and Fuzz Testing}, 
      author={Ana Nunez and Nafis Tanveer Islam and Sumit Kumar Jha and Peyman Najafirad},
      year={2024},
      eprint={2409.10737},
      archivePrefix={arXiv},
      primaryClass={cs.SE},
      url={https://arxiv.org/abs/2409.10737}, 
}

@misc{Platzer,
author = {Platzer, André},
year = {2019},
month = {10},
pages = {},
title = {Overview of Logical Foundations of Cyber-Physical Systems},
doi = {10.48550/arXiv.1910.11232},
archivePrefix={arXiv},
primaryClass= {cs.LO},
url={https://arxiv.org/abs/1910.11232v1},
}

@InProceedings{horizon2,
author="Del Giudice, Nicola
and Loreti, Michele
and Quadrini, Michela
and Rehman, Aniqa",
editor="Margaria, Tiziana
and Steffen, Bernhard",
title="Monitoring Local and Global Properties of Collective Adaptive Systems",
booktitle="Leveraging Applications of Formal Methods, Verification and Validation. Rigorous Engineering of Collective Adaptive Systems",
year="2025",
publisher="Springer Nature Switzerland",
address="Cham",
pages="281--296",
}

@InProceedings{Muccini,
author="Vaidhyanathan, Karthik
and Muccini, Henry",
editor="Bianculli, Domenico
and Sartaj, Hassan
and Andrikopoulos, Vasilios
and Pautasso, Cesare
and Mikkonen, Tommi
and Perez, Jennifer
and Bure{\v{s}}, Tom{\'a}{\v{s}}
and De Sanctis, Martina
and Muccini, Henry
and Navarro, Elena
and Soliman, Mohamed
and Zdun, Uwe",
title="Software Architecture in the Age of Agentic AI",
booktitle="Software Architecture. ECSA 2025 Tracks and Workshops",
year="2026",
publisher="Springer Nature Switzerland",
address="Cham",
pages="41--49",
isbn="978-3-032-04403-7"
}

@INPROCEEDINGS{architsmells,

  author={Pandini, Gabriele and Martini, Antonio and Videsjorden, Adela Nedisan and Fontana, Francesca Arcelli},

  booktitle={2025 IEEE 22nd International Conference on Software Architecture Companion (ICSA-C)}, 

  title={An Exploratory Study on Architectural Smell Refactoring Using Large Languages Models}, 

  year={2025},

  volume={},

  number={},

  pages={462-471},

  doi={10.1109/ICSA-C65153.2025.00070}}

\end{document}